# AN APPROACH TO REDUCE COMPUTATIONAL LOAD: PRECALCULATING GAIN MATRICES FOR AN LQR CONTROLLER OF A FOUR-AXIS MANIPULATOR USING STATE SPACE KINEMATICS

ALISTAIR KEILLER

ABSTRACT. When designing a power or CPU constrained device where a four-axis robotic arm is required and access to the Robot Operating System (ROS) is not an option, finding an efficient state space controller for a four-axis arm can be an obstacle. In this paper, I explore a method to optimize the computing power required for a computer algebra system (CAS) to compute linear quadratic regulator (LQR) matrices by precomputing the gain matrix for different states. Example C++ code is provided on Github, along with ideas for further exploration.

## 1. Introduction

This paper aims to outline the mathematics behind a state space controller for a four-jointed arm. The CAS code for generating the final equation is generalized to $N$ joints, building on the work of J. Doyle et al. [1] and A. Sinha et al. [2]. To achieve this, there are four analytical steps: first, we find the forward kinematics of the arm, which determines the location of each joint based on the angles of the arm; then, the task at hand involves determining the angles of the arm based on the desired location of the grabber; next, the energy of the arm, including inertial, potential, and kinetic energy, should be considered; finally, a state space controller can be derived from these energies using Euler Langrange Mechanics.

## 2. Forward Kinematics

Figure 1 shows the name and location of each arm joint in 3D space:

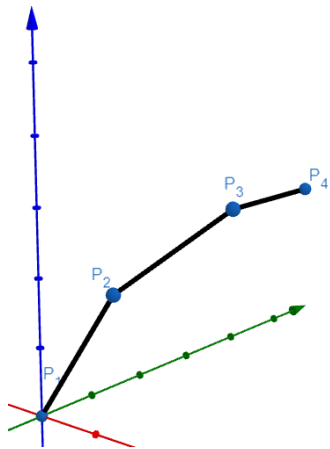

FIGURE 1. Conceptual illustration of the four joints of the arm locations in three-dimensional space

To simplify the understanding of this arm, let's separate the three joints that rotate in the same plane from the joint that rotates around the $Z$/vertical axis. By focusing first on the two-dimensional plane, where most of the complexity lies, we can then transform back into the three-dimensional plane. This eliminates repetition and redundancy in the analysis.

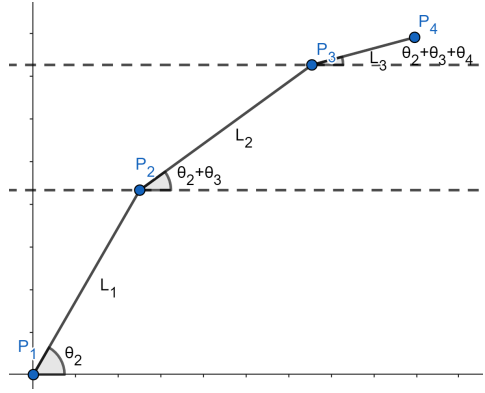

FIGURE 2. Reduced from three dimensions, a two-dimensional plane of arm joints

From Figure 2, we can see that in this two-dimensional plane
$$P_1 = (0,0)$$
$$P_2 = P_1 + (L_1 \sin(\theta_2), L_1 \cos(\theta_2))$$
$$P_3 = P_2 + (L_2 \sin(\theta_2 + \theta_3), L_2 \cos(\theta_2 + \theta_3))$$
$$P_4 = P_3 + (L_3 \sin(\theta_2 + \theta_3 + \theta_4), L_3 \cos(\theta_2 + \theta_3 + \theta_4))$$
and this expands to
$$P_1 = (0,0)$$
$$P_2 = (L_1 \sin(\theta_2), L_1 \cos(\theta_2))$$
$$P_3 = (L_1 \sin(\theta_2) + L_2 \sin(\theta_2 + \theta_3), L_1 \cos(\theta_2) + L_2 \cos(\theta_2 + \theta_3))$$
$$P_4 = (L_1 \sin(\theta_2) + L_2 \sin(\theta_2 + \theta_3) + L_3 \sin(\theta_2 + \theta_3 + \theta_4), L_1 \cos(\theta_2) + L_2 \cos(\theta_2 + \theta_3) + L_3 \cos(\theta_2 + \theta_3 + \theta_4)).$$

Then to convert the 2D coordinates in the joint plane to 3D real coordinates given the rotation of the rotating base ( $\theta_1$ ), we take
$$f(P) = \left(P_x \sin(\theta_1), P_x \cos(\theta_2), P_y\right).$$

Next, by plugging-in our two-dimensional coordinates into the three-dimensional transformation gives us our desired result: three-dimensional coordinates.

### 3. Inverse Kinematics (IK)

First, we start with the equation for the geometrical IK for a two-jointed arm. As Uzmay et al. shows [3], the resulting equations are:

$$q_2 = -\arccos\left(\frac{x^2 + y^2 - L_1^2 - L_2^2}{2L_1 L_2}\right)$$

$$q_1 = \arctan\left(\frac{y}{x}\right) - \arctan\left(\frac{L_2 \sin(q_2)}{L_1 + L_2 \cos(q_3)}\right)$$

for Figure 3:

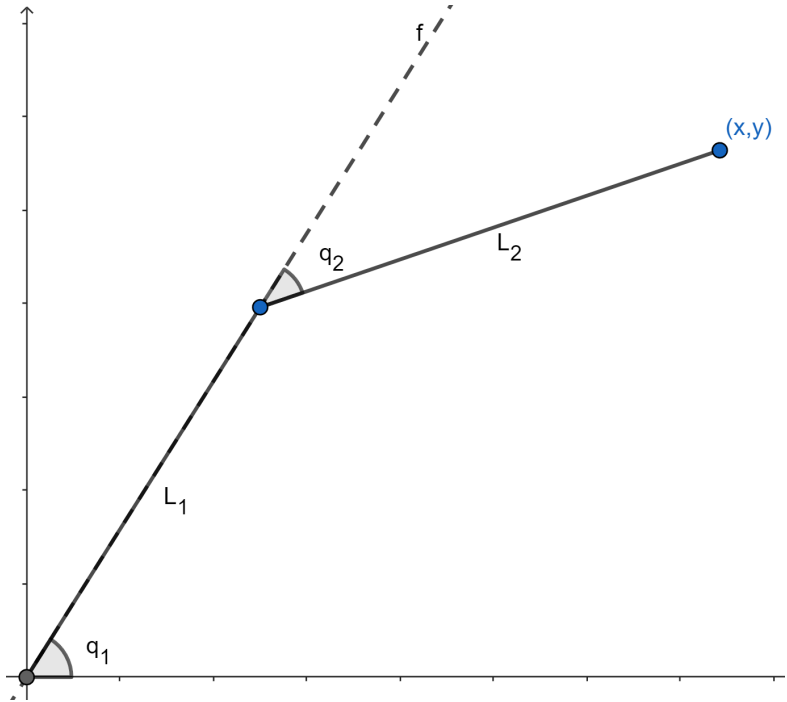

Figure 3. a two-jointed arm

To apply this to our four-axis problem, let's assume that the manipulator is flat. This means that to get the end of the manipulator to point to $(x, y)$, we need to get the start of the manipulator to point to $(x - L_3, y)$, when looking at the two-dimensional plane of the joints. Therefore, by examining the joint plane, we can perform a two-dimensional inverse kinematics calculation to determine the required values of $\theta_2$ and $\theta_3$ to position the start of the manipulator at $(x - L_3, y)$. Next, we determine $\theta_4$ to ensure that the manipulator is level, and $\theta_1$ to ensure that the joint plane intersects the desired point. The final equations are as follows:

$$\theta_1 = \arctan\left(\frac{x}{y}\right)$$

$$\theta_3 = -\arccos\left(\frac{x^2 + y^2 + z^2 - L_1^2 - L_2^2}{2L_1 L_2}\right)$$

$$\theta_2 = \arctan\left(\frac{z}{\sqrt{x^2 + y^2}}\right) - \arctan\left(\frac{L_2 \sin(q_2)}{L_1 + L_2 \cos(q_3)}\right)$$

$$\theta_4 = -\theta_2 - \theta_3.$$

## 4. Inertia of a Line Segment

The objective of this task is to determine the inertia of a line segment rotating around apoint R. The line segment is defined by two points, $P_A = (x_1, y_1)$ to $P_B = (x_2, y_2)$, as shown in Figure 4:

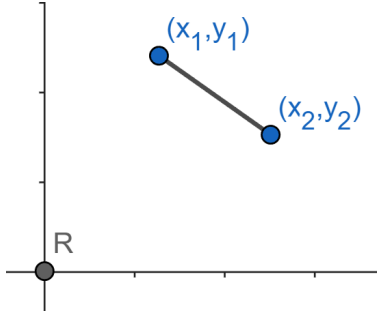

Figure 4. A Line Segment Rotating Around $R$

We can parametrize this segment over $t = [0, 1]$ with
$$f(t) = (x_1 + (x_2 - x_1)t, y_1 + (y_2 - y_1)t).$$

We want to find
$$I(P_A, P_B, m) = \int r^2 dm = m \int_0^1 r^2 dt = m \int_0^1 x^2 + y^2 dt = m \int_0^1 (x_1 + (x_2 - x_1)t)^2 + (y_1 + (y_2 - y_1)t)^2 dt$$
$$= \frac{m}{3}(x_1^2 + x_1 x_2 + x_2^2 + y_1^2 + y_1 y_2 + y_2^2).$$

## 5. Potential Energy of The Arm

### 5.1. Potential Energy of Point Masses. 
The potential as a result of the point masses is
$$\sum mgh = g\left(m_1\left(P_{1_y} - P_{1_y}\right) + m_2\left(P_{2_y} - P_{1_y}\right) + m_3\left(P_{3_y} - P_{1_y}\right) + m_4\left(P_{4_y} - P_{1_y}\right)\right).$$
$$= g\left(m_2 P_{2_y} + m_3 P_{3_y} + m_4 P_{4_y}\right).$$

5.2. **Potential Energy of Line Segments.** The gravitational potential energy can be calculated as the average of the lower and upper heights of the line segments, due to the constant mass distribution of the line segment in the vertical frame and the linear gravitational potential energy. The potential energy is

$$\sum Mg\frac{h_1+h_2}{2} = g\left(M_1\frac{P_{1_y}+P_{2_y}}{2} + M_2\frac{P_{2_y}+P_{3_y}}{2} + M_3\frac{P_{3_y}+P_{4_y}}{2}\right).$$

## 6. Kinetic Energy of Arm

To determine the arm's kinetic energy, we must first calculate its rotational inertia around each joint.

6.1. **Calculating $J_3$.**

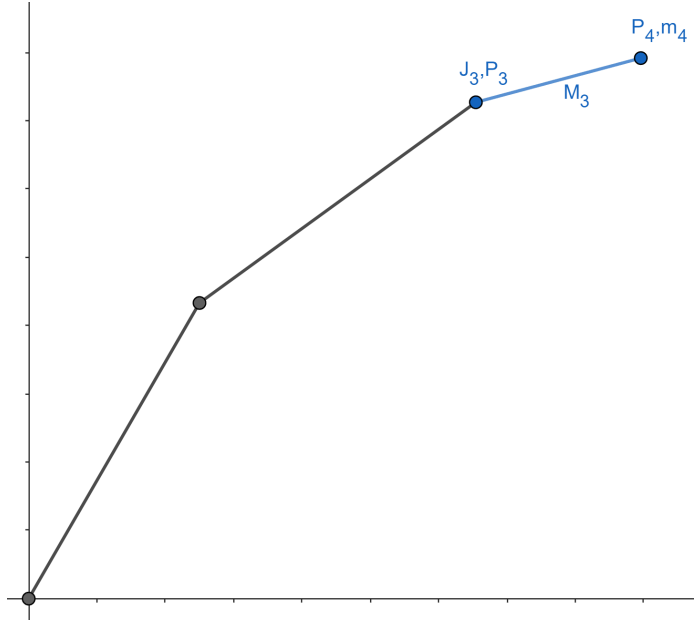

Figure 5. The Arm Focused on J3

Figure 5 shows the useful values for calculating the inertia around $J_3$. To calculate the inertia for each linear mass, we apply the formula for the inertia of a line segment. For each point mass, we use the formula $I(P,m) = mr^2 = m(x^2+y^2)$. In this case, we have a total of 1 line segment with inertia $I(P_3-P_3, P_4-P_3, M_3)$ and 1 point mass with inertia $I(P_4-P_3, m_4)$. This expands to total inertia of $m_4L_3^2 + \frac{1}{3}M_3L_3^2$. Next, to find the kinetic energy, we plug this into the rotational kinetic energy equation, which is $KE = \frac{1}{2}I\omega^2 = \frac{1}{2}(m_4L_3^2 + \frac{1}{3}M_3L_3^2)\theta_{4'}^2$.

6.2. **Calculating $J_2$.** Similarly, for $J_2$, the inertia will be

$$I(P_2-P_2, P_3-P_2, M_2) + I(P_3-P_2, P_4-P_2, M_3)$$

for the line segments and

$$I(P_3 - P_2, m_3) + I(P_4 - P_2, m_4)$$

for the point masses. This means the total inertia will be

$$I(P_2 - P_2, P_3 - P_2, M_2) + I(P_3 - P_2, P_4 - P_2, M_3) + I(P_3 - P_2, m_3) + I(P_4 - P_2, m_4)$$

and the kinetic energy will be

$$\frac{1}{2}(I(P_2 - P_2, P_3 - P_2, M_2) + I(P_3 - P_2, P_4 - P_2, M_3) + I(P_3 - P_2, m_3) + I(P_4 - P_2, m_4))\theta_{3'}^2.$$

6.3. **Calculating $J_1$.** Finally, for $J_1$ the inertia will be

$$I(P_1 - P_1, P_2 - P_1, M_1) + I(P_2 - P_1, P_3 - P_1, M_2) + I(P_3 - P_1, P_4 - P_1, M_3) +$$
$$I(P_2 - P_1, m_2) + I(P_3 - P_1, m_3) + I(P_4 - P_1, m_4).$$

As a result, the kinetic energy will be

$$\frac{1}{2}(I(P_1 - P_1, P_2 - P_1, M_1) + I(P_2 - P_1, P_3 - P_1, M_2) + I(P_3 - P_1, P_4 - P_1, M_3) +$$
$$I(P_2 - P_1, m_2) + I(P_3 - P_1, m_3) + I(P_4 - P_1, m_4))\theta_{2'}^2.$$

## 7. Euler Lagrange Equation

The Lagrangian can be calculated using the potential and kinetic energies found previously, following the approach demonstrated by Inigo et al. [4]. The Lagrangian is $KE - PE$. To get $\alpha = \theta''$, we need to solve the Euler Lagrange equation $\frac{\partial f}{\partial \theta} = \frac{d}{dt}\frac{\partial f}{\partial \theta'}$. By performing this process for each joint, a system of equations is obtained that can be solved to determine the value of $\theta'' = [\theta_{1''}, \theta_{2''}, \theta_{3''}, \theta_{4''}]$, We are aware of the evolution of the motorless system. To consider the motor torque, simply include the term $\theta'' + = \frac{\tau}{I}$. Finally, we have the full equation for $\theta''$, which is now our simulation of the arm.

## 8. State Space

To build an LQR controller to find the optimal path in this simulation, we need to compute the $A$ and $B$ matrices as input to the WPIlib LQR function. For this, we need to establish state and input vectors. Our state will be $x = [\theta, \theta']\}$, while our input will be $u = \tau$. $A = \text{jacobian}\left(\frac{d}{dt}x, x\right) = \text{jacobian}([\theta', \theta'']], x)$, while $B = \text{jacobian}([\theta', \theta''], u)$. We can plug in our solution for $\theta''$ as a function of $\theta$, $\theta'$, and $\tau$ to get $A$ and $B$ in terms of $x$ and $u$ or $\theta$, $\theta'$, and $\tau$.

8.1. **Fixing State Space.** Now that the state space is computed, we still run into a problem of circular reasoning: $\tau$ is an input to the LQR controller. However, the entire purpose of our state space model and the LQR controller is to compute the optimal $\tau$ to get to the desired location. In an affine system, the $\tau$ would cancel out in the calculation of $A$ and $B$, but I have shown in an informal proof that this is not possible for our four-joint arm. So, as a non-affine system, we are left with circular reasoning. One solution is to cache the $\tau$ from the last step of the LQR controller call and use it as an approximate $\tau$ to input to the next LQR.

8.2. **Precalculating values to improve performance.** As Frigerio et al. [5], a problem with this approach is that the final resulting equations for $A$ and $B$ are extremely slow. Despite optimizations and approximations, as well as precomputing $\theta' = 0$ for our desired state, the resulting C++ code is still 47kb and consists mostly of trigonometric functions. This, combined with the slow computation of the LQR controller, requires significant computational power. To reduce the computational load, we can precompute the gain matrix for different states. However, the size of the lookup table for an 8-dimensional space grows with $n^8$,, which is not scalable. A possible solution is to implement an algorithm inspired by mesh algorithms used in CAD programs to efficiently store the precomputed matrices.

## 9. Code on Github

All equations were computed using the Symbolics.jl CAS [6] to compile the precomputed matrices into C++ code. The code was then tested in a simulated GUI built with Makie.jl [7] in the Julia programming language. The corresponding code is available here:

https://github.com/AlistairKeiller/RobotIK

Stanford University
*Email address:* akeiller@stanford.edu
*URL:* https://github.com/AlistairKeiller/RobotIK